\documentclass{article}

\usepackage[utf8]{inputenc} 
\usepackage[T1]{fontenc}    
\usepackage{hyperref}       
\usepackage{url}            
\usepackage{booktabs}       
\usepackage{amsfonts}       
\usepackage{nicefrac}       
\usepackage{microtype}      
\usepackage{graphicx}
\usepackage{gensymb}

\usepackage{geometry}
\usepackage{authblk}

\author[1]{Jing Huang}
\author[1]{Viswanath Sivakumar}
\author[1,2]{Mher Mnatsakanyan}
\author[1]{Guan Pang}
\affil[1]{Facebook Inc.}
\affil[2]{University of California, Berkeley}

\begin{document}

\title{Improving Rotated Text Detection with Rotation Region Proposal Networks}
\maketitle


\begin{abstract}
A significant number of images shared on social media platforms such as Facebook and Instagram contain text in various forms. It's increasingly becoming commonplace for bad actors to share misinformation, hate speech or other kinds of harmful content as text overlaid on images on such platforms. A scene-text understanding system should hence be able to handle text in various orientations that the adversary might use. Moreover, such a system can be incorporated into screen readers used to aid the visually impaired. In this work, we extend the scene-text extraction system at Facebook, Rosetta \cite{borisyuk2018rosetta}, to efficiently handle text in various orientations. Specifically, we incorporate the Rotation Region Proposal Networks (RRPN) \cite{ma2018arbitrary} in our text extraction pipeline and offer practical suggestions for building and deploying a model for detecting and recognizing text in arbitrary orientations efficiently. Experimental results show a significant improvement on detecting rotated text.
\end{abstract}

\section{Introduction}

Understanding text in images shared on platforms such as Facebook and Instagram along with the context in which it appears makes it possible to proactively identify inappropriate or harmful content and keep our community safe. While over the years we had gotten good at handling policy-violating text composed in posts and captions, we were exposed to hate speech, clickbait, policy-violating ads, and other low quality content that manifested as part of an image. Motivated by this problem, Rosetta \cite{borisyuk2018rosetta} was built to extract overlaid and scene-text from images and video frames and identify policy violating content. Rosetta extracts text from more than a billion public Facebook and Instagram images and video frames (in a wide variety of languages), daily and in real time, and feeds the output to upstream classifiers to understand the context of the text and the image together.

Rosetta employs a two-step approach. The text detection model, based on Faster-RCNN \cite{ren2015faster} with the ResNet convolutional body replaced with a ShuffleNet-based \cite{zhang1707shufflenet} architecture, is responsible of detecting regions of the image that contain words. Each detected region is cropped and fed to a fully-convolutional character-based text recognition model to extract the word. Widely used object detection models such as Faster-RCNN are designed for generic cases and thus output rectangular bounding boxes. For scene-text extraction, this is insufficient as text may come in arbitrary orientations while the second-stage text recognition model typically expects an image patch with horizontal text. According to an error analysis on Rosetta soon after it was deployed, oriented text was found to be the most common source of mistakes with orientations as minimal as 20\degree failing to be recognized correctly. Therefore, it's important to enable the system to correctly handle rotated text amongst all the adversarial cases we might face.

\begin{figure}
    \centering
    \includegraphics[width=0.95\linewidth]{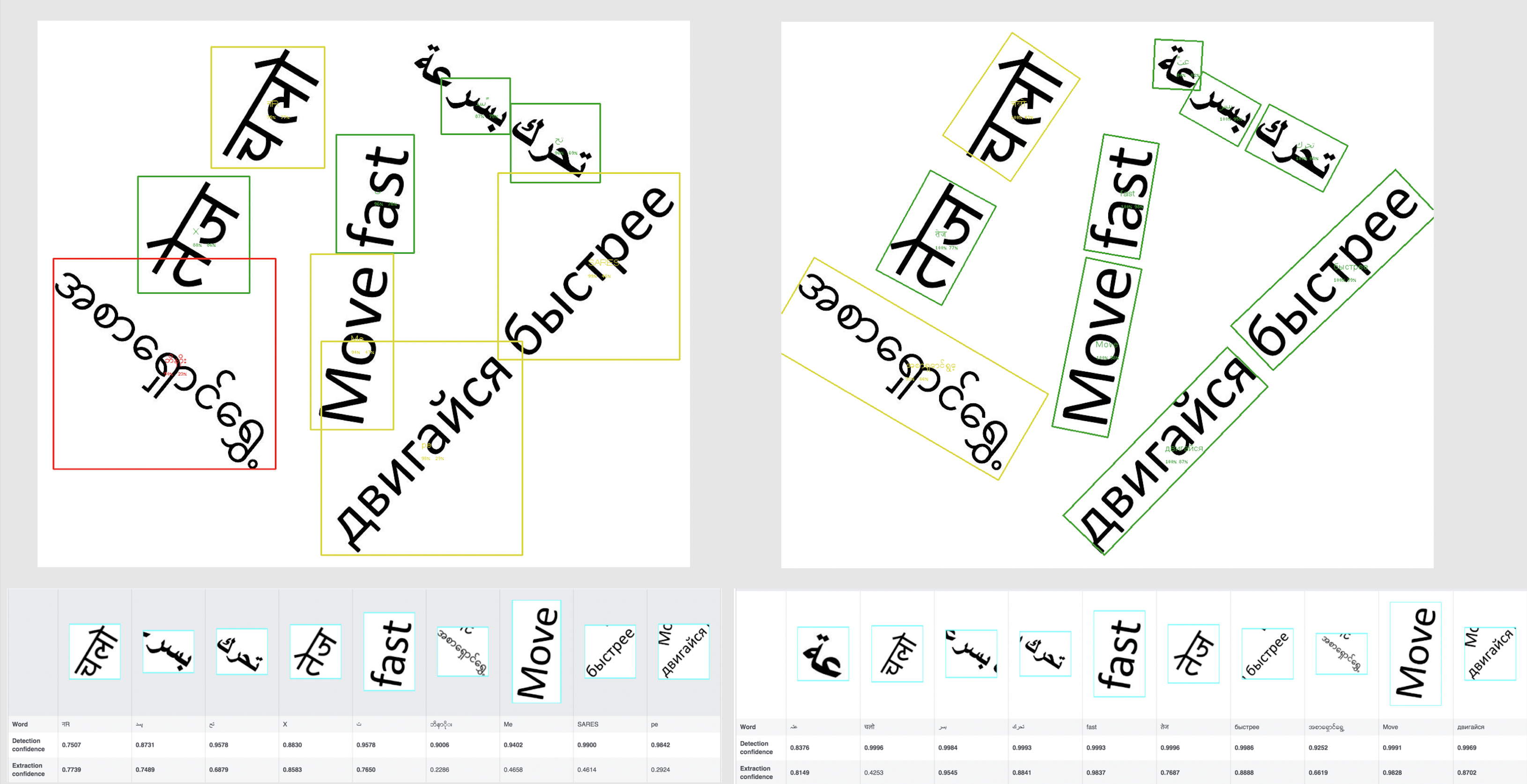}
    \caption{Comparison of text extraction between non-rotated model and rotated models.}
    \label{fig:multilingual_prod_rrpn_comparison}
    \vspace{-3mm}
\end{figure}

In traditional Faster-RCNN detection, while most words are detected with a bounding box, we cannot correctly infer the actual words due to lack of textual orientation information (Figure \ref{fig:multilingual_prod_rrpn_comparison}). One solution is to apply a trained Spatial Transformer Network on the rectangular patch, which benefits from being standalone module. Another approach is to predict an oriented bounding box in detection stage, which benefits from being able to train end-to-end. The two approaches can be combined together if needed. We follow the idea of \cite{ma2018arbitrary} to replace the Region Proposal Network (RPN) in our Faster-RCNN-based detection pipeline with RRPN. There are a few differences in our implementation compared to original RRPN: \newline
\textbf{RRoI transformation:} We use Rotated Region of Interest Align (RRoI-Align) that applies bi-linear interpolation, instead of Rotated Region of Interest Pooling (RRoI-Pooling) as RoI transformation to avoid misaligned result on the boundaries due to rounding. \newline
\textbf{Boundary breaking anchors:} The original RRPN will filter out boundary breaking R-anchors during training and they had to use a border padding of 0.25 times each side to reserve more positive proposals. Our approach automatically performs on-demand padding in the RRoI-Align operator as well as bounding box transformation between the detection and recognition pipeline. \newline
\textbf{Orientation coverage:} Due to trade-offs between orientation coverage and computational efficiency, the original RRPN crops the angle range to be within $[-45, 135)$. However, it's not a symmetrical range as the text rotated by $(-90, -45)$ degrees would be treated as text rotated by $(90, 135)$ degrees. Therefore, we use a more natural orientation coverage of $(-90, 90)$ degrees so that any text that is not rotated by more than 90 degrees can be correctly identified. The RPN anchors are chosen accordingly.

\section{Rotational Region Proposal Network}

\subsection{Rotated box Representation}
Traditionally, a non-rotated bounding box is represented as $(x, y, w, h)$ where $(x, y)$ is the top-left corner of the box, or $(x_1, y_1, x_2, y_2)$, where $(x_1, y_1)$ is the top-left corner and $(x_2, y_2)$ is the bottom-right corner of the box. RRPN \cite{ma2018arbitrary} uses $(x_c, y_c, h, w, \theta)$ as the representation, where $(x_c, y_c)$ represents the geometric center of the bounding box. The orientation $\theta$ is the angle from the positive direction of x-axis to the direction parallel to the width of the rotated box. We follow a similar representation.

\subsection{Rotated Anchors}
To fit the objects to different sizes, RPN uses two parameters to control the size and shape of anchors, i.e., scale and aspect ratio. For rotated text detection, anchor generation should include orientation as well. The original RRPN uses $(8, 16, 32)$ as scale anchors (essentially it's equivalent to $(128, 256, 512)$ in our pipeline with a scaling multiplier of $16$), $(0.125, 0.2, 0.5)$ as aspect ratio anchors, and $(-30, 0, 30, 60, 90, 120)$ as angle anchors. Other than the trade-off between orientation coverage and computational efficiency leading to using half of the angle space (180 degrees), it didn't give an explanation for the specific cropping angle of $(-45, 135)$ degrees, which means boxes in range of $(-90, -45)$ degrees would be treated as text rotated by $(90, 135)$ degrees. From theoretical and empirical perspective, we believe that it's because the convergence of angles would become unstable around the cropping angles and they try to place the boundary at some non-right-angle degrees. On the other hand, in our latest experiments we use the same scales and aspect ratios as our non-rotated model (5 scale anchors $(32, 64, 128, 256, 512)$, 7 aspect ratio anchors $(0.03125, 0.0625, 0.125, 0.25, 0.5, 1, 2)$), while the angle anchors are between $-90$ and $+90$ degrees: $(-90, -60, -30, 0, 30, 60, 90)$. While the range of $(-90, 90)$ degrees is used for now, our pipeline also supports detecting arbitrary angles (covering all 360 degrees) with a config change. The extra dimension for angle anchors multiplies the number of anchors by $7$ and thus increases the memory usage during training by a non-negligible amount, which we handle by using in-place operators.

\subsection{Rotated RoI Transformation}
Given the region proposals, we'd like to obtain a fixed-size feature map (e.g. $7 \times 7$). Operators like RoI Pooling allows us to achieve this goal. In the RRPN paper \cite{ma2018arbitrary}, the authors extended RoI Pooling to RRoI Pooling by applying rotation transformation. On the other hand, quantization in RoI and RRoI pooling creates misaligned result on the boundaries. To handle this, we implemented Rotated RoI Align operator that applies bi-linear interpolation.

\section{Proximity Estimation for Rotated Box}
Assuming we have two proposals, how should we evaluate which one is closer to the ground truth? This is the key question for proposal selection as well as non-maximum suppression (NMS). For horizontal boxes, it's sufficient to use IoU (intersection over union) between two boxes. However, IoU alone is not enough to estimate the proximity of rotated boxes. Given a horizontal box, we can have up to $4$ ways of representing it: a box rotated by $0\degree, 90\degree, 180\degree, 270\degree$. They occupy exactly the same pixels of the image, but they mean different things in OCR when processed by the text-recognition model. For example, the letter '$V$' might be recognized as '$<$' if the predicted box has $90\degree$ angle, or '$>$' if the box has $-90\degree$ angle. Therefore, angle difference between the rotated boxes needs to be taken into account. We apply the same strategy as \cite{ma2018arbitrary}: positive R-anchors are those with the highest IoU overlap or an IoU larger than $0.7$ with respect to the ground truth, and an intersection angle with respect to the ground truth of less than $30$ degrees. Negative R-anchors are those with an IoU lower than $0.3$, or an IoU larger than $0.7$ but with an intersection angle with a ground truth larger than $30$ degrees. Regions not belonging to either positive or negative are not used during training.


\subsection{Training}
While the key concept is not complicated, the change to support the 5-dimensional bounding box representation without breaking the current pipeline turned out to be pretty invasive. Besides the core changes mentioned above, various efforts were made in supporting training with mixed non-rotated and rotated datasets to bias the model towards the predominant scenario of non-rotated text without sacrificing too much accuracy, on-the-fly rotated data augmentation, and supporting evaluation of rotated boxes.

\subsection{Inference}
For efficient inference, we perform int8 quantization on the model to reduce runtime and memory-bandwidth usage. Also, we implemented efficient CPU versions of Caffe2 operators to handle rotated boxes including RotatedRoIAlign, BBoxTransform, Non-Maximum Suppression, GenerateProposals and BoxWithNMSLimit, leading to no increase in inference time compared to baseline non-rotated model. These Caffe2 operators have been open-sourced at https://github.com/pytorch/pytorch/blob/master/caffe2/operators.

\begin{figure}
    \centering
    \includegraphics[width=0.95\linewidth]{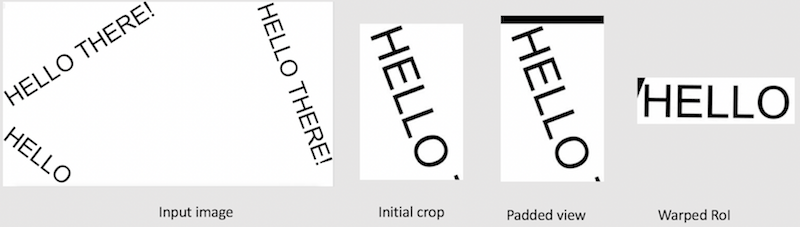}
    \caption{Bounding box tranformation on predicted rotated boxes between detection and recognition.}
    \label{fig:rbbox_transform}
\end{figure}

Finally, we need to perform rotated box transformation in order to generate the image patch based on predicted rotated box parameters and feed it to the recognition pipeline. Traditional Faster-RCNN clips any boxes that might overflow image boundaries, but doing so for rotated boxes isn't straight-forward and would lead to cutting-out some characters since the detection output is still a rectangle as opposed to a polygon. To handle this, we pad the original image until the boxes don't overflow the boundaries anymore \ref{fig:rbbox_transform} in the following steps: First, we use the horizontal bounding rectangle of the predicted rotated box to crop out the region from the original image. Then, we add zero-padding so that the center of the image patch is the same as the predicted rotated box. Finally, we use warp-affine transformation in OpenCV to simultaneously rotate the patch and crop out the region of interest with the correct width and height from the prediction.

\section{Experiments}
\subsection{Qualitative Results}

Using the RRPN approach, we can correctly predict the rotated bounding box for the example in the beginning (the image and text detection and recognition result in the right of Figure \ref{fig:multilingual_prod_rrpn_comparison}).


\subsection{Quantitive Evaluation}

\begin{figure}
    \centering
    \includegraphics[width=0.99\linewidth]{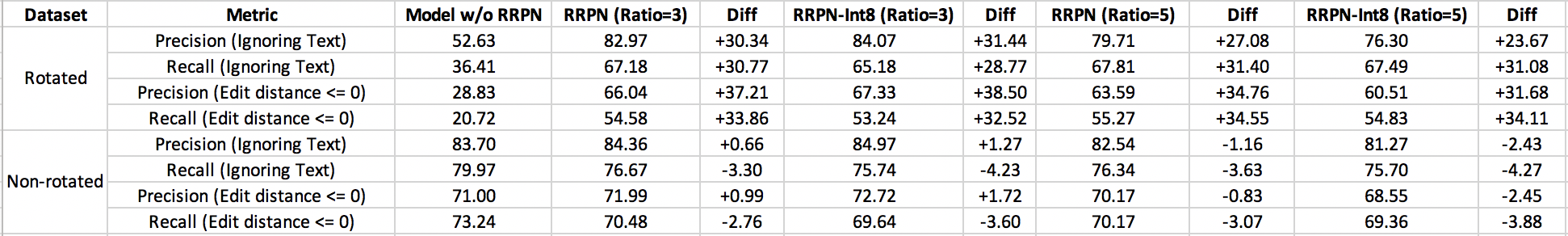}
    \caption{Quantitative evaluation.}
    \label{fig:quantative_evaluation}
    \vspace{-3mm}
\end{figure}

We performed end-to-end evaluation on two datasets, Rotated (with uniform random rotation in $[-90, 90]$ degrees) and Non-rotated datasets, on five models. (1) Baseline model without RRPN; (2) RRPN ($Ratio=3$) trained on an $3:1$ rotation-augmented multilingual dataset; (3) RRPN ($Ratio=3$) model after int8 quantization; (4) RRPN ($Ratio=5$) trained on $5:1$ rotation-augmented multilingual dataset; (5) RRPN ($Ratio=5$) model after int8 quantization. All of them are trained for around $675k$ iterations. Results are summarized in Figure \ref{fig:quantative_evaluation}. The RRPN models perform significantly better on Rotated dataset, while slightly worse on Non-rotated dataset. It would be more interesting to train for as many iterations as possible (for example, RRPN with $ratio=3$ should be trained for $>2M$ iterations for it to see as many non-rotated examples as the baseline non-rotated model) and evaluate the results. If the result shows significant trade-off between the models, it might be an evidence of hitting the model capacity. We also compared the latency of int8-quantized rotated model with the baseline non-rotated model during inference and found no significant increase, which means we can process images at similar throughput as before.

\section{Conclusions}
We adapted Rotation Region Proposal Network for detecting oriented text, and made important improvements based on it. This further improves the quality Rosetta, the OCR system at Facebook, to protect against adversarial text in images including engagement bait, policy violating ads and images with profanity and hate speech, as well as improves the accuracy of screen readers to make Facebook more accessible for the visually impaired. In the future, this framework could be extended to more adversarial scenarios such as text with general affine transformations.

\section*{Acknowledgement}
The authors would like to thank Albert Gordo, Peizhao Zhang, Manohar Paluri, Tyler Matthews, Mahalia Miller and others who contributed, supported and collaborated with us during the development and deployment of our system.

{\small
\bibliographystyle{ieee}
\bibliography{egbib}

\begin{thebibliography}{1}\itemsep=-1pt

\bibitem{borisyuk2018rosetta}
F.~Borisyuk, A.~Gordo, and V.~Sivakumar.
\newblock Rosetta: Large scale system for text detection and recognition in
  images.
\newblock In {\em Proceedings of the 24th ACM SIGKDD International Conference
  on Knowledge Discovery \& Data Mining}, pages 71--79. ACM, 2018.

\bibitem{ma2018arbitrary}
J.~Ma, W.~Shao, H.~Ye, L.~Wang, H.~Wang, Y.~Zheng, and X.~Xue.
\newblock Arbitrary-oriented scene text detection via rotation proposals.
\newblock {\em IEEE Transactions on Multimedia}, 2018.

\bibitem{ren2015faster}
S.~Ren, K.~He, R.~Girshick, and J.~Sun.
\newblock Faster r-cnn: Towards real-time object detection with region proposal
  networks.
\newblock In {\em Advances in neural information processing systems}, pages
  91--99, 2015.

\bibitem{zhang1707shufflenet}
X.~Zhang, X.~Zhou, M.~Lin, and J.~Sun.
\newblock Shufflenet: An extremely efficient convolutional neural network for
  mobile devices. arxiv 2017.
\newblock {\em arXiv preprint arXiv:1707.01083}, 2017.

\end{thebibliography}
}

\end{document}